\documentclass[10pt,letterpaper]{article} 

\usepackage{url}

\usepackage{graphicx}

\usepackage{amsmath,amssymb}
\def\D{\mathrm{d}}

\usepackage{times}

\graphicspath{{.}{pics/}}

\usepackage[accepted]{icml2016}

\icmltitlerunning{Minimum Regret Search for Single- and Multi-Task Optimization}

\begin{document}

\twocolumn[
\icmltitle{Minimum Regret Search for Single- and Multi-Task Optimization}

\icmlauthor{Jan Hendrik Metzen}{janmetzen@mailbox.org}
\icmladdress{Universit{\"a}t Bremen, 28359 Bremen, Germany \\
			 Corporate Research, Robert Bosch GmbH, 70442 Stuttgart, Germany}

\icmlkeywords{Bayesian Optimization, Contextual Policy Search, Multi-Task Learning, Gaussian Process}

\vskip 0.3in
]


\begin{abstract}
We propose minimum regret search (MRS), a novel acquisition function for Bayesian
optimization. MRS bears similarities with information-theoretic approaches
such as entropy search (ES). However, while ES aims in each query at maximizing the
information gain with respect to the global maximum, MRS aims at minimizing the
expected simple regret of its ultimate recommendation for the optimum. While empirically ES and MRS perform similar in most of the
cases, MRS produces fewer outliers with high simple regret than ES. We provide empirical
results both for a synthetic single-task optimization problem as well as for a
simulated multi-task robotic control problem.
\end{abstract}

\section{INTRODUCTION}

Bayesian optimization \citep[BO,][]{shahriari_taking_2016} denotes a sequential,
model-based, global approach for optimizing black-box functions. It is particularly well-suited for problems
which are non-convex, do not necessarily provide derivatives, are expensive to evaluate (either
computationally, economically, or morally), and can potentially be noisy. Under
these conditions, there is typically no guarantee for finding the true optimum of the
function with a finite number of function evaluations. Instead, one often aims
at finding a solution which has small \emph{simple regret} \cite{bubeck_pure_2009} with regard to the true optimum, where 
simple regret denotes the difference of the true optimal function value 
and the function value of the ``solution'' selected by the algorithm after a finite number of function evaluations. BO
aims at finding such a solution of small simple regret while  minimizing at the same
time the number of evaluations of the expensive target function. For this, BO
maintains a probabilistic surrogate model of the objective function, and a
myopic utility or acquisition function, which defines the ``usefulness'' of
performing an additional function evaluation at a certain input for learning about the optimum.

BO has been applied to a diverse set of problems, ranging from hyperparameter optimization
of machine learning models \cite{bergstra_algorithms_2011,snoek_practical_2012} over robotics
\cite{calandra_bayesian_2015,lizotte_automatic_2007,kroemer_combining_2010,alonso_an_2015} to
sensor networks \cite{srinivas_gaussian_2010} and environmental monitoring
\cite{6385653}. For a more comprehensive overview of application areas, we refer
to \citet{shahriari_taking_2016}.

A critical component for the performance of BO is the acquisition function,
which controls the exploratory behavior of the sequential search procedure.
Different kinds of acquisition functions have been proposed, ranging from
improvement-based acquisition functions over optimistic acquisition functions to
information- theoretic acquisition functions (see Section
\ref{Section:Background}). In the latter class, the group of entropy
search-based approaches
\cite{villemonteix_informational_2008, hennig_entropy_2012,hernandez-lobato_predictive_2014}, 
which aims at maximizing the information gain regarding
the true optimum, has achieved state-of-the-art performance on a number of
synthetic and real-world problems. However, performance is often reported as the
median over many runs, which bears the risk that the median masks ``outlier'' runs that
perform considerably worse than the rest. In fact, our results indicate that
the performance of sampling-based entropy search is not necessarily better
than traditional and cheaper acquisition functions according to the mean simple regret.

In this work, we propose minimum regret search (MRS), a novel acquisition
function that explicitly aims at minimizing the expected simple regret
(Section \ref{Section:MRS}). MRS performs well according to both the mean and
median performance on a synthetic problem (Section
\ref{Section:ResultsSingleTask}). Moreover, we discuss how MRS can be extended
to multi-task optimization problems (Section \ref{Section:CMRS}) and present
empirical results on a simulated robotic control problem (Section
\ref{Section:ResultsMultiTask}).

\section{BACKGROUND} \label{Section:Background}

In this section, we provide a brief overview of
\textbf{Bayesian Optimization} (BO). We refer to \citet{shahriari_taking_2016} for a recent more
extensive review of BO. BO can be applied to black-box optimization
problems, which can be framed as optimizing an objective function $f:
\mathcal{X} \to \mathbb{R}$ on some bounded set $\mathcal X \subset
\mathbb{R}^D$. In contrast to most  other black-box optimization methods, BO is
a global method which makes use of all previous evaluations of $f(\mathbf{x})$
rather than using only a subset of the history for approximating a local
gradient or Hessian. For this, BO maintains a \emph{probabilistic model} for
$f(\mathbf{x})$, typically a Gaussian process \citep[GP,][]{rasmussen_gaussian_2006}, and uses this model for deciding
at which $\mathbf{x}_{n+1}$ the function $f$ will be evaluated next.

Assume we have already queried $n$ datapoints and observed their (noisy) function values $\mathcal{D}_n=\{(\mathbf{x}_i,
y_i)\}_{i=1}^n$. The choice of the next query point $\mathbf{x}_{n+1}$ is based
on a utility function over the GP posterior, the so-called \emph{acquisition
function} $a: \mathcal X \to \mathbb{R}$, via $\mathbf{x}_{n+1}=
\arg\max_\mathbf{x} a(\mathbf{x})$. Since the maximum of the acquisition
function cannot be computed directly, a global optimizer is typically
used to determine $\mathbf{x}_{n+1}$. A common strategy
is to use
DIRECT \citep{jones_lipschitzian_1993} to find the approximate global maximum,
followed by L-BFGS \citep{byrd_limited-memory_1995} to refine it.

The first class of acquisition functions are \emph{optimistic policies} such as
the upper confidence bound (UCB) acquisition function. $a_{UCB}$ aims at minimizing the regret during the course of BO and has the form $$a_{UCB}(\mathbf{x};
\mathcal{D}_n) = \mu_{n}(\mathbf{x}) + \kappa_n \sigma_{n}(\mathbf{x}),$$ where
$\kappa_n$ is a tunable parameter which balances exploitation ($\kappa_n = 0$) and
exploration ($\kappa_n \gg 0$), and $\mu_{n}$ and $\sigma_{n}$ denote mean and
standard deviation of the GP at $\mathbf{x}$, respectively.
\citet{srinivas_gaussian_2010} proposed GP-UCB, which entails a specific schedule for $\kappa_n$ that yields provable cumulative regret bounds.

The second class of acquisition functions are \emph{improvement-based policies}, such as the probability of improvement \citep[PI,][]{kushner_new_1964} over the current best value, which can be calculated in closed-form for a GP model:
$$a_{PI}(\mathbf{x}; \mathcal{D}_n) := \mathbb{P}[f(\mathbf{x}) > \tau] = \Phi(\gamma(\mathbf{x})),$$
where $\gamma(x) = (\mu_n(\mathbf{x}) - \tau) / \sigma_n(x)$, $\Phi(\cdot)$ denotes the cumulative distribution function of the
standard Gaussian, and $\tau$ denotes the incumbent, typically the best function value observed so far: $\tau = \max_i y_i$.
Since PI exploits quite aggressively \cite{jones_taxonomy_2001}, a more popular
alternative is the \emph{expected improvement} \citep[EI,][]{mockus_application_1978} over the
current best value $ a_{EI}(\mathbf{x}; \mathcal{D}) := \mathbb{E}[(f(x) - \tau)\mathbb{I}(f(x) > \tau)]$, which can again be computed in closed form for a GP model as
$a_{EI}(\mathbf{x}; \mathcal{D})
  = \sigma_n(\mathbf{x})\left(\gamma(\mathbf{x})\Phi(\gamma(\mathbf{x})) + \phi(\gamma(\mathbf{x}))\right)$,
where $\phi(\cdot)$ denotes the standard Gaussian density function.  A
generalization of EI is the \emph{knowledge gradient} factor \cite
{frazier_knowledge-gradient_2009}, which can better handle noisy observations,
which impede the estimation of the incumbent. The knowledge gradient requires
defining a set $A_n$ from which one would choose the final solution.

The third class of acquisition functions are \emph{information-based policies},
which entail Thompson sampling and entropy search
\citep[ES,][]{villemonteix_informational_2008, hennig_entropy_2012,hernandez-lobato_predictive_2014}.
Let $p^\star(x \vert \mathcal{D}_n)$ denote the
posterior distribution of the unknown optimizer $\mathbf{x}^\star = \arg\max_{\mathbf{x} \in \mathcal{X}} f(\mathbf{x})$  after
observing $\mathcal{D}_n$. The objective of ES is to select the query point that
results in the maximal reduction in the differential entropy of $p^\star$. More
formally, the entropy search acquisition function is defined as
$$ a_{ES}(\mathbf{x}, \mathcal{D}_n)
  = H(\mathbf{x}^\star \vert \mathcal{D}_n) - \mathbb{E}_{y \vert \mathbf{x}, \mathcal{D}_n}
	[H(\mathbf{x}^\star \vert \mathcal{D}_n \cup \{(\mathbf{x}, y))],$$
where $H(\mathbf{x}^\star \vert \mathcal{D}_n)$ denotes the differential entropy
of $p^\star(x \vert \mathcal{D}_n)$ and the expectation is with respect to the
predictive distribution of the GP at $\mathbf{x}$, which is a normal
distribution $y \sim \mathcal{N}(\mu_{n}(\mathbf{x}), \sigma_{n}^2(\mathbf{x}) +
\sigma^2)$. Computing $a_{ES}$ directly is intractable for continuous spaces
$\mathcal{X}$; prior work has discretized $\mathcal{X}$ and used either Monte
Carlo sampling \cite{villemonteix_informational_2008} or expectation propagation
\cite{hennig_entropy_2012}. While the former may require many Monte Carlo
samples to reduce variance, the latter incurs a run time that is quartic in the
number of representer points used in the discretization of $p^\star$. An
alternative formulation is obtained by exploiting the symmetric property of the
mutual information, which allows rewriting the ES acquisition function as
$$ a_{PES}(\mathbf{x}, \mathcal{D}_n)
  = H(y \vert \mathcal{D}_n, \mathbf{x}) - \mathbb{E}_{\mathbf{x}^\star \vert \mathcal{D}_n}
	[H(y \vert \mathcal{D}_n, \mathbf{x}, \mathbf{x}^\star)].$$
This acquisition function is known as predictive entropy search
\citep[PES,][]{hernandez-lobato_predictive_2014}. PES does not require discretization
and allows a formal treatment of GP hyperparameters.

\textbf{Contextual Policy Search} (CPS)
denotes a model-free approach to reinforcement learning,
in which the (low-level) policy $\pi_\mathbf{\theta}$ is parametrized by a vector
$\mathbf{\theta}$. The choice of $\mathbf{\theta}$ is governed by an upper-level policy
$\pi_{u}$. For generalizing learned policies to multiple tasks,
the task is characterized
by a context vector $\mathbf{s}$ and the upper-level policy
$\pi_{u}(\mathbf{\theta}\vert \mathbf{s})$ is conditioned on the respective
context. The objective of CPS is to learn the upper-level policy
$\pi_{u}$ such that the expected return $J$ over all contexts is
maximized, where $J = \int_s p(\mathbf{s}) \int_\mathbf{\theta}
\pi_{u}(\mathbf{\theta}\vert \mathbf{s}) R(\mathbf{\theta}, \mathbf{s}) \D{\mathbf{\theta}
} \D{\mathbf{s}}$. Here, $p(\mathbf{s})$ is the distribution over contexts and
$R(\mathbf{\theta}, \mathbf{s})$ is the expected return when executing the low
level policy with parameter $\mathbf{\theta}$ in context $\mathbf{s}$. We refer to
\citet{deisenroth_survey_2013} for a recent overview of (contextual) policy
search approaches in robotics.

\section{MINIMUM REGRET SEARCH} \label{Section:MRS}

Information-based policies for Bayesian optimization such as ES and PES
have performed well empirically. However, as we discuss in Section \ref{Section:MRS_Illustration},
their internal objective of minimizing
the uncertainty in the location of the optimizer $\mathbf{x}^\star$, i.e.,
minimizing the differential entropy of $p^\star$, is actually different (albeit
related) to the common external objective of minimizing the \emph{simple regret} of
$\mathbf{\tilde x}_N$, the recommendation of BO for $\mathbf{x}^\star$ after $N$
trials. We define the simple regret of $\mathbf{\tilde x}_N$ as
$R_f(\mathbf{\tilde  x}_N) = f(\mathbf{x}^\star) - f(\mathbf{\tilde  x}_N) = \max_{\mathbf{x}} f(\mathbf{x}) - f(\mathbf{\tilde x}_N)$.

Clearly, $\mathbf{x}^\star$ has zero and thus minimum simple regret, but a query that
results in the maximal decrease in $H(\mathbf{x}^\star)$ is not necessarily the
one that also results in the maximal decrease in the expected simple regret. 
In this section, we propose \emph{minimum regret search} (MRS), 
which explicitly aims at minimizing the expected simple
regret.

\subsection{Formulation}
Let $\mathcal{X} \subset \mathcal{R}^D$ be some bounded domain and $f: \mathcal{X} \mapsto
\mathbb{R}$ be a function. We are interested in finding the maximum
$\mathbf{x}^\star$ of $f$ on $\mathcal{X}$. Let there be a probability measure
$p(f)$ over the space of functions $f: \mathcal{X} \mapsto \mathbb{R}$, such as
a GP. Based on this $p(f)$, we would ultimately like to
select an $\mathbf{\tilde x}$ which has minimum simple regret
$R_f(\mathbf{\tilde x})$. We define the expected simple regret ER of
selecting parameter $\mathbf{x}$ under $p(f)$ as:
$$\text{ER}(p)(\mathbf{x}) = \mathbb{E}_{p(f)}[R_f(\mathbf{x})]
= \mathbb{E}_{p(f)}[\max_{\mathbf{x}} f(\mathbf{x}) - f(\mathbf{
x})]$$

In $N$-step Bayesian optimization, we are given a budget of $N$ function evaluations and are to choose a sequence of $N$ query points
$X^q_N=\{\mathbf{x}^q_1, \dots, \mathbf{x}^q_N\}$ for which we evaluate $y_i = f(\mathbf{x}^q_i) +
\epsilon$ to obtain $\mathcal{D}_N = \{\mathbf{x}^q_i, y_i\}_{i=1}^N$. Based on this, 
 $p_N(f) = p(f \vert \mathcal{D}_N)$ is estimated and a point
$\mathbf{\tilde x}_N$ is recommended as the estimate of $\mathbf{x}^\star$ such that the
expected simple regret under $p_N$ is minimized, i.e., $\mathbf{\tilde x}_N = \arg\min_\mathbf{x}
\text{ER}(p_N)(\mathbf{x})$.
The minimizer $\mathbf{\tilde x}_N$ of the expected simple regret under fixed
$p(f)$ can be approximated efficiently in the case of a GP since it is
identical to the maximizer of the GP's mean $\mu_N(\mathbf{x})$. 

However, in general, $p(f)$ depends on data and it is desirable to select the data such
that $\text{ER}$ is also minimized with regard to the resulting $p(f)$.
We are thus interested in choosing a
sequence $X^q_N$ such that we minimize the expected simple regret of $\mathbf{\tilde x}_N$ with respect to the $p_N$, where $p_N$ depends on $X^q_N$ and the (potentially noisy) observations $y_i$.
As choosing the optimal sequence $X^q_N$ at once is intractable, we follow the
common approach in BO and select $\mathbf{x}^q_n$
sequentially in a myopic way. However, as $\mathbf{\tilde x}_N$ itself depends
on $\mathcal{D}_N$ and is thus unknown for $n < N$, we have to use proxies for it based on the currently
available subset $\mathcal{D}_n \subset \mathcal{D}_N$.  One simple choice for a proxy is to use
the point which has minimal expected simple regret under $p_n(f) =
p(f \vert \mathcal{D}_n)$, i.e.,
$\mathbf{\tilde x} = \arg\min_\mathbf{x} \text{ER}(p_n)(\mathbf{x})$.
Let us denote the updated
probability measure on $f$ after performing a query at $\mathbf{x}^q$ and
observing the function value $y=f(\mathbf{x}^q) + \epsilon$ by
$p^{[\mathbf{x}^q, y]}_n = p(f\vert \mathcal{D}_n \cup \{\mathbf{x}^q, y\})$.
We define the acquisition function MRS$^{\text{point}}$ as the expected reduction of the minimum expected regret for a query at
$\mathbf{x}^q$, i.e.,
\begin{equation*}
\begin{split}
a_{\text{MRS}^{\text{point}}}(\mathbf{x}^q)
	& = \min_{\mathbf{\tilde x}}\text{ER}(p_n)(\mathbf{\tilde x}) \\
	 &- \mathbb{E}_{y \vert p_n(f), \mathbf{x}^q}[\min_{\mathbf{\tilde x}}  \text{ER}(p^{[\mathbf{x}^q, y]}_n)(\mathbf{\tilde x})],
\end{split}
\end{equation*}
where the expectation is with
respect to $p_n(f)$'s predictive distribution at $\mathbf{x}^q$ and we drop the
implicit dependence on $p_n(f)$ in the notation of $a_{\text{MRS}^{\text{point}}}$.
The next query point $\mathbf{x}^q_{n+1}$ would thus be selected as the maximizer of
$a_{\text{MRS}^{\text{point}}}$.

One potential drawback of MRS$^{\text{point}}$, however,  is that it does not account for
the inherent uncertainty about $\mathbf{\tilde x}_N$. To address this shortcoming, we propose using
the measure $p^\star_{\mathcal{D}_n} = p^\star(\mathbf{x} \vert \mathcal{D}_n)$ as defined in
entropy search (see Section \ref{Section:Background}) as proxy for $\mathbf{\tilde x}_N$.
We denote the resulting acquisition function by MRS and define it analogously to $a_{\text{MRS}^{\text{point}}}$:
\begin{equation*}
\begin{split}
a_{\text{MRS}}(\mathbf{x}^q)
   & = \mathbb{E}_{\mathbf{\tilde x} \sim p^\star_n}[\text{ER}(p_n)(\mathbf{\tilde x})] \\
	  & - \mathbb{E}_{y \vert p_n(f), \mathbf{x}^q}[
		  \mathbb{E}_{\mathbf{\tilde x} \sim p^\star_{\mathcal{D}_n \cup \{(\mathbf{x}^q, y)\}}}[
			 \text{ER}(p^{[\mathbf{x}^q, y]}_n)(\mathbf{\tilde x})]].
\end{split}
\end{equation*}

MRS can thus be seen as a more Bayesian treatment, where we marginalize our
uncertainty about $\mathbf{\tilde x}_N$, while MRS$^\text{point}$ is more akin
to a point-estimate since we use a single point (the minimizer of the expected simple regret)
as proxy for $\mathbf{\tilde x}_N$.

\subsection{Approximation}
Since several quantities in MRS cannot be computed in closed form, we
resort to similar discretizations and approximations as proposed for entropy search by \citet{hennig_entropy_2012}. We
focus here on sampling based approximations; for an alternative way of approximating $\mathbb{E}_{p(f)}$ based on
expectation propagation, we refer to \citet{hennig_entropy_2012}.

\begin{figure*}
\centering
\includegraphics{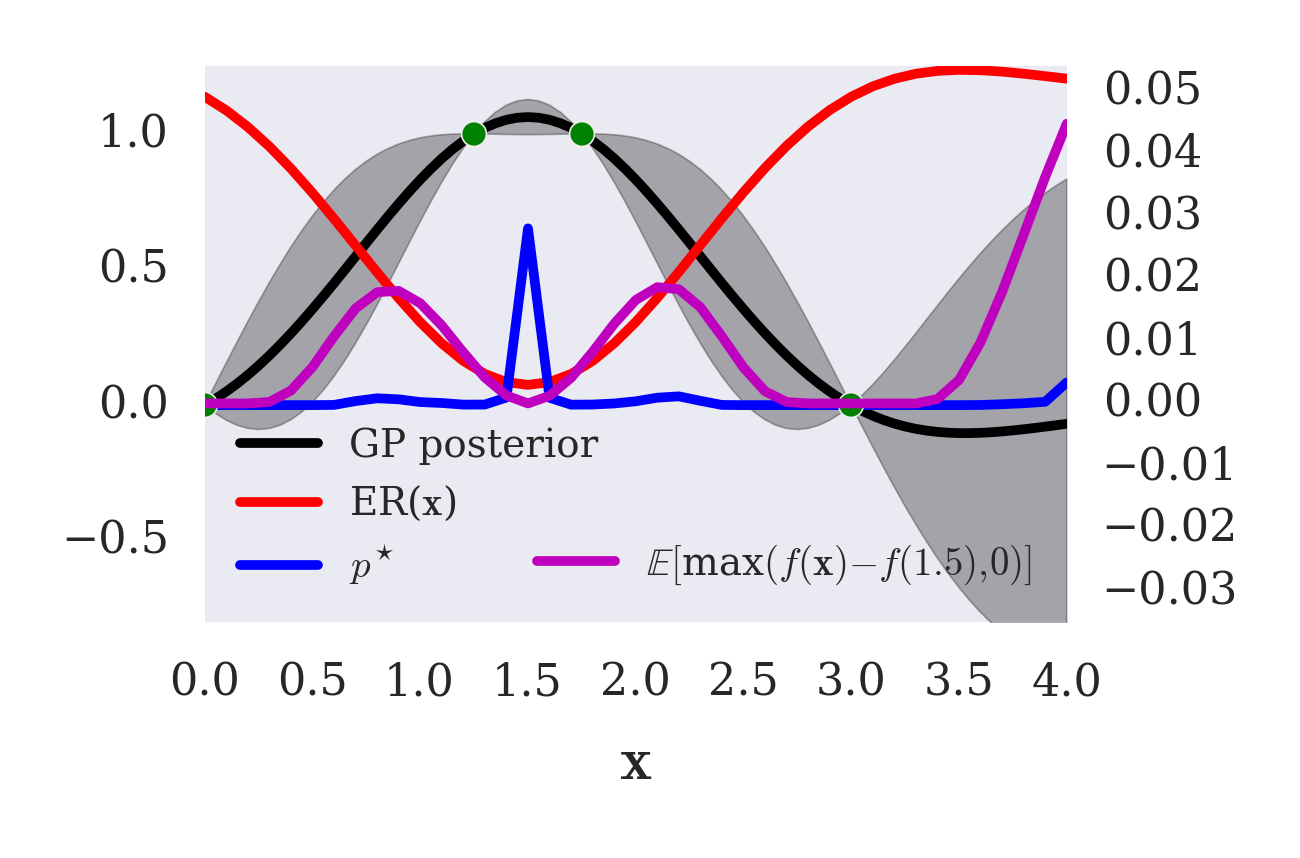}
\includegraphics{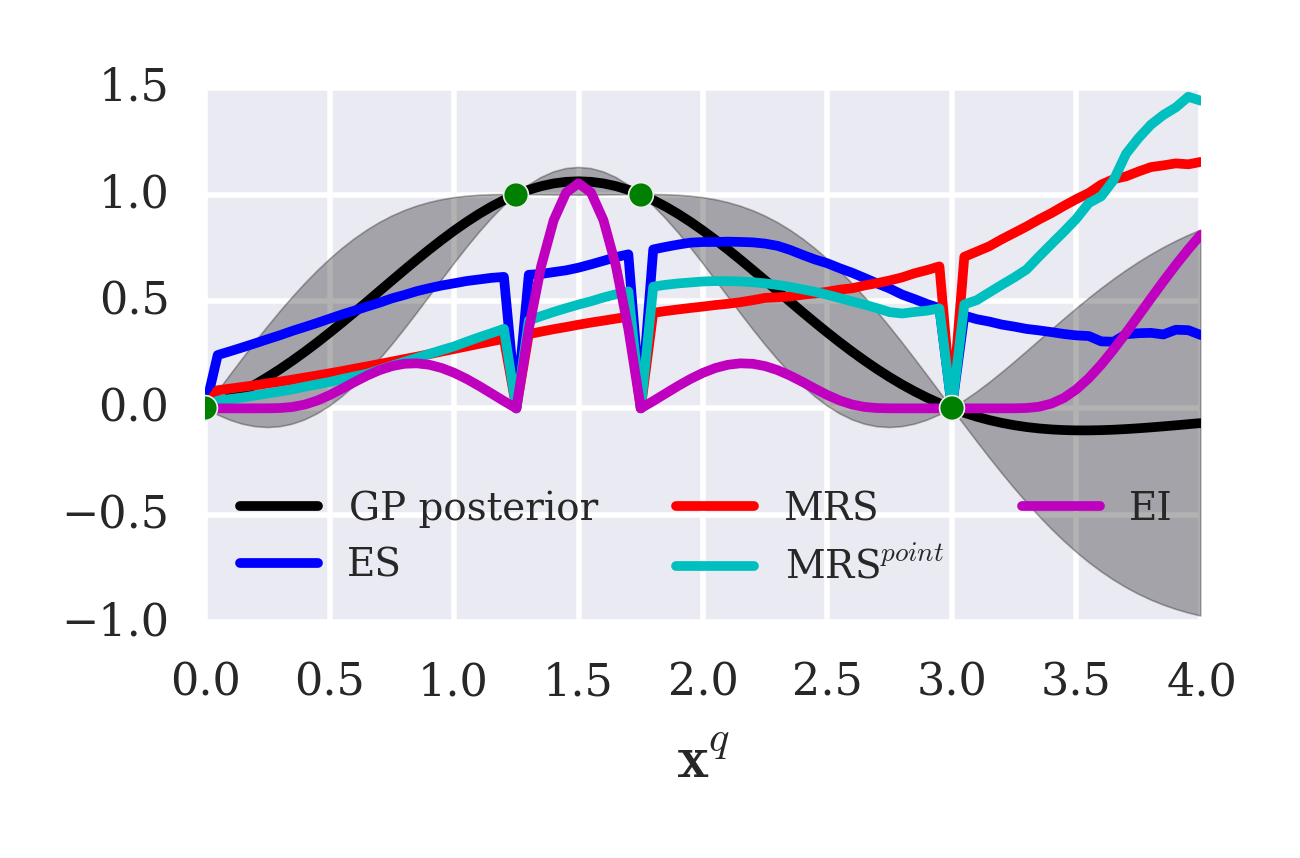}
\caption{(Left) Illustration of GP posterior, probability of maximum $p^\star$, expected regret ER, and  $\mathbb{E}[\max(f(\mathbf{x}) - f(1.5), 0)]$ (scale on the right-hand side). (Right) Illustration of GP posterior and different acquisition function. Absolute values have been normalized such that the mean value of an acquisition function is $0.5$. Best seen in color.}
\label{fig:MRS_illustration}
\end{figure*}

Firstly, we approximate $\mathbb{E}_{p(f)}$ by taking $n_f$ Monte Carlo samples
from $p(f)$, which is straightforward in the case of GPs.
Secondly, we approximate $\mathbb{E}_{y \vert p(f), \mathbf{x}^q}$  by taking
$n_y$ Monte Carlo samples from $p(f)$'s predictive distribution at
$\mathbf{x}_q$. And thirdly, we discretize $p^\star$ to a finite set of $n_r$
representer points chosen from a non-uniform measure, which turns
$\mathbb{E}_{\mathbf{\tilde x} \sim p^\star_n}$ in the definition of
$\text{MRS}(\mathbf{x}^q)$ into a weighted sum. The discretization of $p^\star$ is discussed by
\citet{hennig_entropy_2012} in detail; we select the representer points as
follows: for each representer point, we sample $250$ candidate points uniform
randomly from $\mathcal{X}$ and select the representer point by Thompson
sampling from $p(f)$ on the candidate points. Moreover, estimating $p^\star_{\mathcal{D}_n}$ 
on the representer points can be done relatively cheap by reusing the samples used 
for approximating $\mathbb{E}_{p(f)}$, which incurs a small bias which had, however, a negligible 
effect in preliminary experiments.

The resulting estimate of $a_{\text{MRS}}$ would have high variance and would require $n_f$ to be
chosen relatively large; however, we can reduce the variance considerably by using
common random numbers \citep{kahn_methods_1953} in the estimation of
$\text{ER}(p)(\mathbf{x})$ for different $p(f)$.

\subsection{Illustration} \label{Section:MRS_Illustration}

Figure \ref{fig:MRS_illustration} presents an illustration of different
acquisition functions on a simple one-dimensional target function. The left
graphic shows a hypothetical GP posterior (illustrated by its mean and standard
deviation) for length scale $l=0.75$, and the resulting probability of $\mathbf{x}$
being the optimum of $f$ denoted by $p^\star$.
Moreover, the expected simple regret of selecting $\mathbf{x}$ denoted by
$\text{ER}(\mathbf{x})$ is shown. The minimum of $\text{ER}(\mathbf{x})$ and the maximum
of $p^\star$ are both at $\mathbf{\tilde x}=1.5$. The expected regret of $\mathbf{\tilde x}$
is approximately $\text{ER}(\mathbf{\tilde x}) = 0.07$. We plot $\mathbb{E}[\max(f(\mathbf{x}) -
f(\mathbf{\tilde x}), 0)]$ additionally to shed some light onto situations in which $
\mathbf{\tilde x} = 1.5$ would incur a significant regret: this quantity shows that most of the
expected regret of $\mathbf{\tilde x}$ stems from situations where the ``true''
optimum is located at $\mathbf{x} \geq 3.5$. This can be explained by the
observation that this area has high uncertainty and is at the same time largely
uncorrelated with $\mathbf{\tilde x} = 1.5$ because of the small length-scale of the GP.

The right graphic compares different acquisition functions for $n_f = 1000$,
$n_r=41$, $n_y = 11$, and fixed representer points.  Since the assumed GP is noise-free, the
acquisition-value of any parameter that has already been evaluated is approximately $0$.  The acquisition functions differ considerably in
their global shape: EI becomes large for areas with close-to-maximal predicted mean or with high uncertainty.
 ES becomes large for parameters which are informative with regard to most of the
probability mass of $p^\star$, i.e., for $\mathbf{x}^q \in [0.5, 3.0]$.  In contrast
MRS$^\text{point}$ becomes maximal for $\mathbf{x}^q \approx 4.0$. This can be
explained as follows: according to the current GP posterior, MRS$^\text{point}$ selects $\mathbf{\tilde x} = 1.5$. As shown in the Figure \ref{fig:MRS_illustration} (left),
most of the expected regret for this value of $\mathbf{\tilde x}$ stems from scenarios
where the true optimum would be at $\mathbf{x} \geq 3.5$. Thus, sampling in this
parameter range can reduce the expected regret considerably---either by confirming that the true value of $f(\mathbf{x})$ on $\mathbf{x} \geq 3.5$ is actually as small as expected or by switching $\mathbf{\tilde x}_{n+1}$ to this area if $f(\mathbf{x})$ turns out to be large. The maximum of MRS
is similar to MRS$^\text{point}$. However, since it also takes the whole measure
$p^\star$ into account, its acquisition surface is smoother in general; in particular, it assigns a larger value to regions such as $\mathbf{x}^q
\approx 3.0$, which do not cause regret for $\mathbf{\tilde x} = 1.5$ but for
alternative choices such as $\mathbf{\tilde x} = 4.0$.

Why does ES not assign a large acquisition value to query points $\mathbf{x}^q \geq 3.0$?
This is because ES does not take into account the correlation of different
(representer) points under $p(f)$. This, however, would be desirable as, e.g.,
reducing uncertainty regarding optimality among two highly correlated points
with large $p^\star$ (for instance $\mathbf{x}_1 = 1.5$ and $\mathbf{x}_2 =
1.55$ in the example) will not change the expected regret considerably since
both points will have nearly identical value under all $f \sim p(f)$. On the
other hand, the value of two points which are nearly uncorrelated under $p(f)$ and have non-zero $p^\star$ such as
$\mathbf{x}_1$ and $\mathbf{x}_3 = 4.0$ might differ considerably under
different $f \sim p(f)$ and choosing the wrong one as $\mathbf{\tilde x}$ might
cause considerable regret. Thus, identifying which of the two is actually
better would reduce the regret considerably. This is exactly why MRS assigns large value to $\mathbf{x}^q \approx 4.0$.

\section{MULTI-TASK MINIMUM REGRET SEARCH} \label{Section:CMRS}

Several extensions of Bayesian optimization for multi-task learning have been
proposed, both for discrete set of tasks \citep{krause_contextual_2011}
and for continuous set of tasks
\cite{metzen_active_2015}. Multi-task BO has been demonstrated to learn
efficiently about a set of discrete tasks concurrently
\cite{krause_contextual_2011}, to allow transferring knowledge learned on
cheaper tasks to more expensive tasks \cite{swersky_multi-task_2013}, and to
yield state-of-the-art performance on low-dimensional contextual policy search
problems \cite{metzen_bayesian_2015}, in particular when combined with active
learning \cite{metzen_active_2015}. In this section, we focus on multi-task BO
for a continuous set of tasks; a similar extension for discrete multi-task
learning would be straightforward.

A continuous set of tasks is encountered for instance when applying BO to
contextual policy search (see Section \ref{Section:Background}). We follow the
formulation of BO-CPS \cite{metzen_bayesian_2015} and adapt it for MRS were
required. In BO-CPS, the set of tasks is encoded in a context vector $\mathbf{s}
\in \mathcal{S}$ and BO-CPS learns a (non-parametric) upper-level policy $\pi_{u}$
which selects for a given context $\mathbf{s}$ the parameters $\mathbf{\theta}
\in \mathcal{X}$ of the low-level policy $\pi_\theta$. The unknown function $f$
corresponds to the expected return $R(\mathbf{\theta}, \mathbf{s})$ of executing
a low-level policy with parameters $\mathbf{\theta}$ in context $\mathbf{s}$.
Thus, the probability measure $p(f)$ (typically a GP) is defined over functions
$f: \mathcal{X} \times \mathcal{S} \mapsto \mathbb{R}$ on the joint
parameter-context space. The probability measure is conditioned on the trials performed so
far, i.e., $p_n(f) = p(f \vert \mathcal{D}_n)$ with $\mathcal{D}_n =
\{((\theta_i, s_i), r_i)\}_{i=1}^n$ for $r_i = R(\mathbf{\theta}_i, \mathbf{s}_i)$. Since $p_n$ is defined over the joint
parameter and context space, experience collected in one context is naturally
generalized to similar contexts. 

In passive BO-CPS on which we focus here (please refer to
\citet{metzen_active_2015} for an active learning approach), the context (task)
$\mathbf{s}_n$ of a trial is determined externally according to $p(\mathbf{s})$, 
for which we assume a uniform distribution in this work.
BO-CPS selects the parameter $\mathbf{\theta}_n$ by conditioning $p(f)$ on
$\mathbf{s}=\mathbf{s}_n$ and finding the maximum of the acquisition function
$a$ for fixed $\mathbf{s}_n$, i.e., $\mathbf{\theta}_n =
\arg\max_\mathbf{\theta} a(\mathbf{\theta}, \mathbf{s}_n)$. Acquisition
functions such as PI and EI are not easily generalized to multi-task problems as they are
defined relative to an incumbent $\tau$, which is typically the best function
value observed so far in a task. Since there are infinitely many tasks and no
task is visited twice with high probability, the notion of an incumbent is not
directly applicable.  In contrast, the acquisition functions GP-UCB
\cite{srinivas_gaussian_2010} and ES \cite{metzen_bayesian_2015} have been extended
straightforwardly and the same approach applies also to MRS.

\section{EXPERMENTS}

\subsection{Synthetic Single-Task Benchmark} \label{Section:ResultsSingleTask}

In the first experiment\footnote{Source code for replicating the reported experiment
is available under \url{https://github.com/jmetzen/bayesian_optimization}.}, 
we conduct a similar analysis as \citet[Section
3.1]{hennig_entropy_2012}: we compare different algorithms on a number of
single-task functions sampled from a generative model, namely from the same GP-based model
that is used by the optimization internally as surrogate model. This precludes
model-mismatch issues and unwanted bias which could be introduced by resorting
to common hand-crafted test functions\footnote{Please refer to the Appendix \ref{section:exp_model_mismatch} for an analogous experiment with model-mismatch.}. More specifically, we choose the
parameter space to be the 2-dimensional unit domain $\mathcal{X} = [0, 1]^2$ and generate test
functions by sampling $250$ function values jointly from a GP with an isotropic RBF kernel
of length scale $l = 0.1$ and unit signal variance. A GP is
fitted to these function values and the resulting posterior mean is used as the
test function. Moreover, Gaussian noise with standard deviation $\sigma =
10^{-3}$ is added to each observation. The GP used as surrogate model in the optimizer employed the same 
isotropic RBF kernel with fixed, identical hyperparameters. 
In order to isolate effects of the different acquisition functions from effects of different
recommendation mechanisms, we used the point which maximizes the GP posterior
mean as recommendation $\mathbf{\tilde x}_N$ regardless of the employed acquisition function.
All algorithms were tested on the same
set of $250$ test functions, and we used $n_f=1000$, $n_r=25$, and $n_y=51$.
We do not provide error-bars on the estimates as the data sets have no parametric distribution. 
However, we provide additional histograms on the regret distribution.

\begin{figure*}
\centering
\includegraphics[width=.9\textwidth]{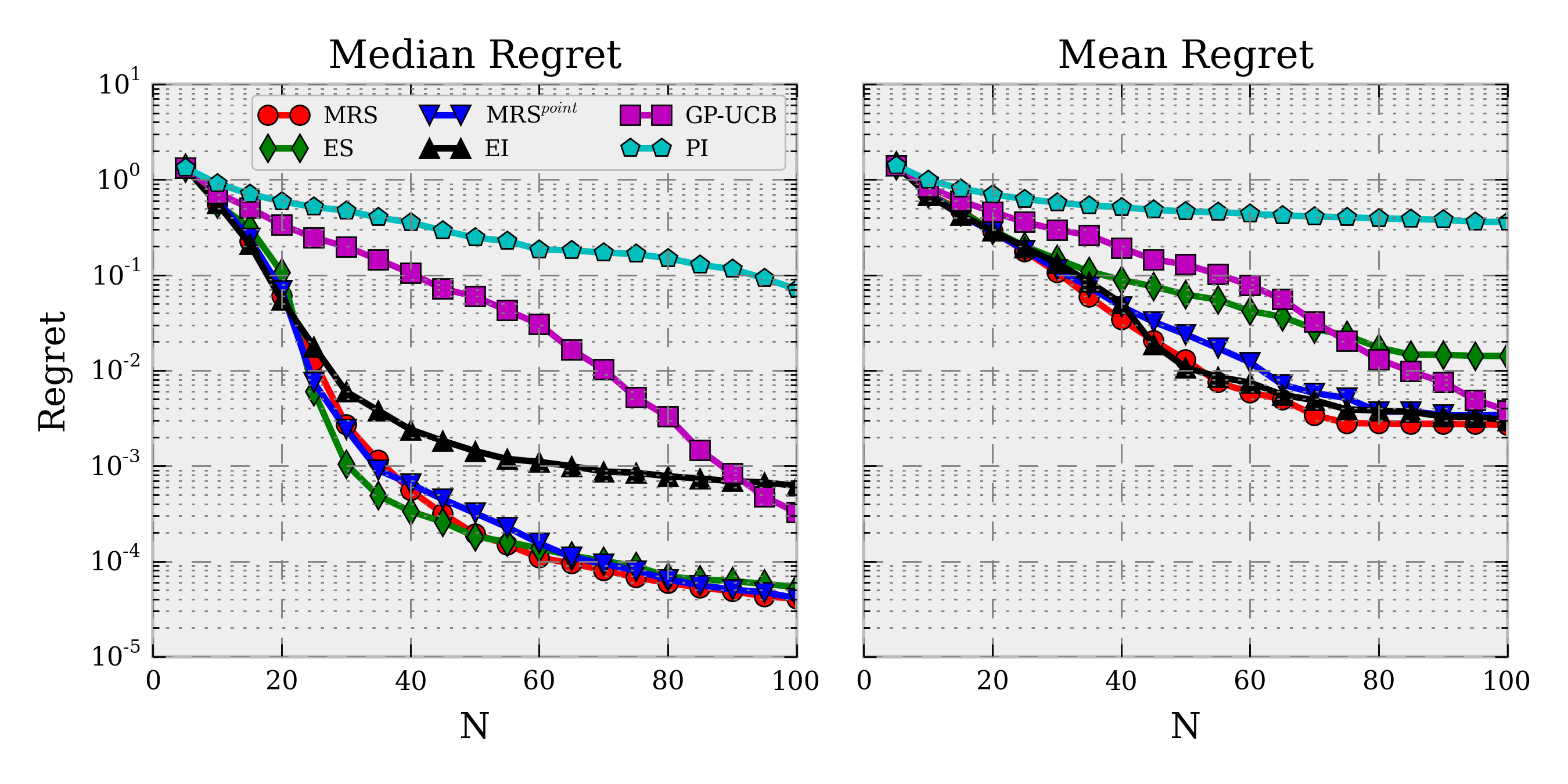}
\includegraphics[width=.9\textwidth]{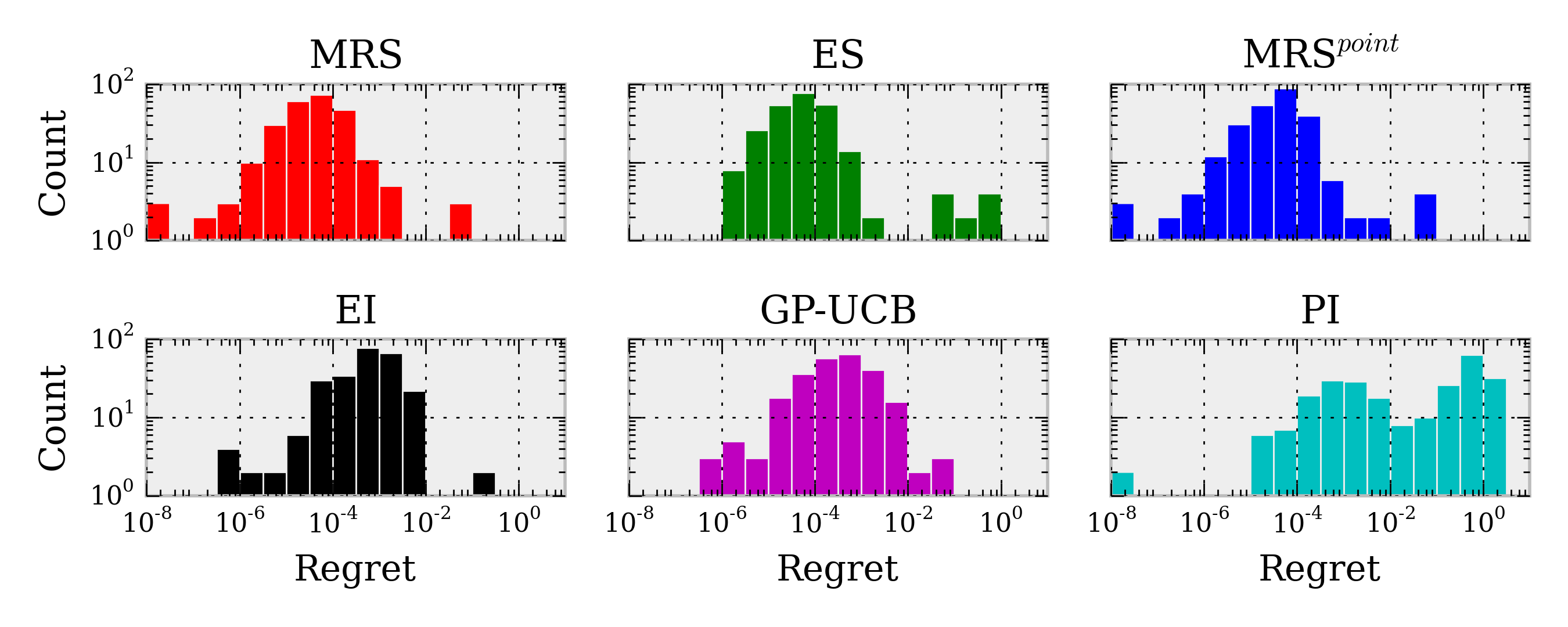}
\caption{(Top) Median and mean simple regret over $250$ repetitions for different acquisition functions. Shown is the simple regret of the recommendation $\mathbf{\tilde x}_N$ after $N$ trials, i.e., the point which maximizes the GP posterior
mean. (Bottom) Histogram of the simple regret after performing $N=100$ trials for different acquisition functions (note the log-scales).}
\label{fig:empirical_comparison}
\end{figure*}

\begin{figure*}
\centering
\includegraphics[width=.9\textwidth]{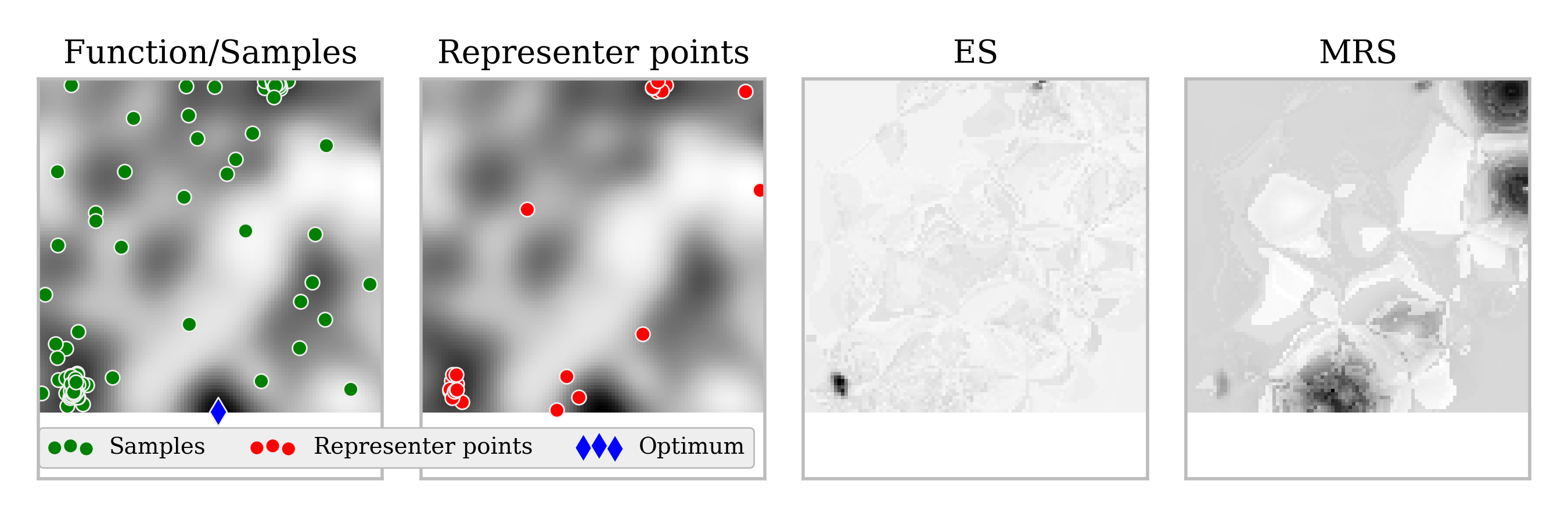}
\caption{Illustration of acquisition functions on a target function for
100 given samples and 25 representer points; darker areas correspond to larger values. ES focuses on sampling in areas with high density of $p^\star$ (many representer points), while MRS focuses on unexplored areas that are populated by representer points (non-zero $p^\star$).
}
\label{fig:es_analysis}
\end{figure*}

Figure \ref{fig:empirical_comparison} summarizes the results for a pure
exploration setting, where we are only interested in the quality of the
algorithm's recommendation for the optimum after $N$ queries but not in the
quality of the queries themselves:  according to the
median of the simple regret (top left), ES, MRS, and MRS$^{point}$ perform
nearly identical, while EI is about an order of magnitude worse. GP-UCB performs
even worse initially but surpasses EI eventually\footnote{GP-UCB would reach 
the same level of mean and median simple regret as MRS eventually after $N=200$ steps 
(in mean and median) with no significant difference  according to a Wilcoxon signed-rank test.}. 
PI performs the worst as it
exploits too aggressively. These results are roughly in correspondence with prior
results \cite{hennig_entropy_2012,hernandez-lobato_predictive_2014} on the same
task; note, however, that \citet{hernandez-lobato_predictive_2014} used a lower
noise level and thus, absolute values are not comparable. However, according to
the \emph{mean} simple regret, the picture changes considerably (top right):
here, MRS, MRS$^{point}$, and EI perform roughly on par while ES is about an order of
magnitude worse. This can be explained by the distribution of the simple
regrets (bottom): while the distributions are fairly non-normal for all
acquisition functions, there are considerably more runs with very high simple
regret ($R_f(\mathbf{\tilde x}_N)>10^{-2}$) for ES (10) than for MRS (4) or EI (4).

We illustrate one such case where ES incurs high simple regret in Figure
\ref{fig:es_analysis}. The same set of representer points has been used for ES and MRS.
While both ES and MRS assign a non-zero density $p^\star$
(representer points) to the area of the true optimum of the function
(bottom center), ES assigns high acquisition value only to areas with a high
density of $p^\star$ in order to further concentrate density in these areas.
Note that this is not due to discretization of $p^\star$ but because of ES'
objective, which is to learn about the precise location of the optimum, irrespective of
how much correlation their is between the representer points according to $p(f)$.
Thus, predictive entropy search \cite{hernandez-lobato_predictive_2014} would likely be affected by the
same deficiency. In contrast, MRS focuses first on areas which have not been
explored and have a non-zero $p^\star$, since those are areas with high expected
simple regret (see Section \ref{Section:MRS_Illustration}). Accordingly, MRS is less likely to incur a high simple regret. In summary, the MRS-based acquisition functions are the only
acquisition functions that perform well both according to the median and the
mean simple regret; moreover, MRS performs slightly better than
MRS$^{point}$ and we will focus on MRS subsequently.

\subsection{Multi-Task Robotic Behavior Learning} \label{Section:ResultsMultiTask}

We present results in the simulated robotic control task used by
\citet{metzen_active_2015}, in which the robot arm COMPI \cite{COMPI} is used to
throw a ball at a target on the ground encoded in a two-dimensional context
vector. The target area is $\mathcal{S} = [1, 2.5]m \times [-1, 1]m$ and the robot arm is
mounted at the origin $(0, 0)$ of this coordinate system. Contexts are sampled uniform randomly from $\mathcal{S}$.
The low-level policy is a joint-space dynamical movement primitives
\citep[DMP,][]{ijspeert_dynamical_2013} with preselected start and goal angle
for each joint and all DMP weights set to 0. This DMP results in throwing a ball
such that it hits the ground close to the center of the target area. Adaptation
to different target positions is achieved by modifying the parameter
$\mathbf{\theta}$: the first component of $\mathbf{\theta}$ corresponds to the
execution time $\tau$ of the DMP, which determines how far the ball is thrown,
and the further components encode the final angle $g_i$ of the $i$-th joint.
We compare the learning performance for different number of controllable joints;
the not-controlled joints keep the preselect goal angles of the initial throw.
The limits on the parameter space are $g_i \in \left[-\frac{\pi}{2}, \frac{\pi}{2}\right]$ and $\tau \in \left[0.4, 2\right]$.

All approaches use a GP with anisotropic Mat{\'e}rn kernel for representing
$p(f)$ and the kernel's length scales and signal variance are selected in each 
BO iteration as point estimates using maximum marginal likelihood. Based on preliminary experiments,
UCB's exploration parameter $\kappa$ is set to a
constant value of $5.0$. For MRS and ES, we use the same parameter values as in
the Section \ref{Section:ResultsSingleTask}, namely $n_f=1000$, $n_r=25$, and $n_y=51$. Moreover, we
add a ``greedy'' acquisition function, which always selects $\theta$ that
maximizes the mean of the GP for the given context (UCB with $\kappa=0$), and a
``random'' acquisition function that selects $\theta$ randomly. The return is
defined as $R(\mathbf{\theta}, \mathbf{s}) = -||\mathbf{s} - b_s||^2 - 0.01
\sum_t v_t^2$, where $b_s$ denotes the position hit by the ball, and $\sum_t
v_t^2$ denotes a penalty term on the sum of squared joint velocities during DMP
execution; both $b_s$ and $\sum_t v_t^2$ depend indirectly on $\mathbf{\theta}$.

\begin{figure*}
\centering
\includegraphics[width=.85\textwidth]{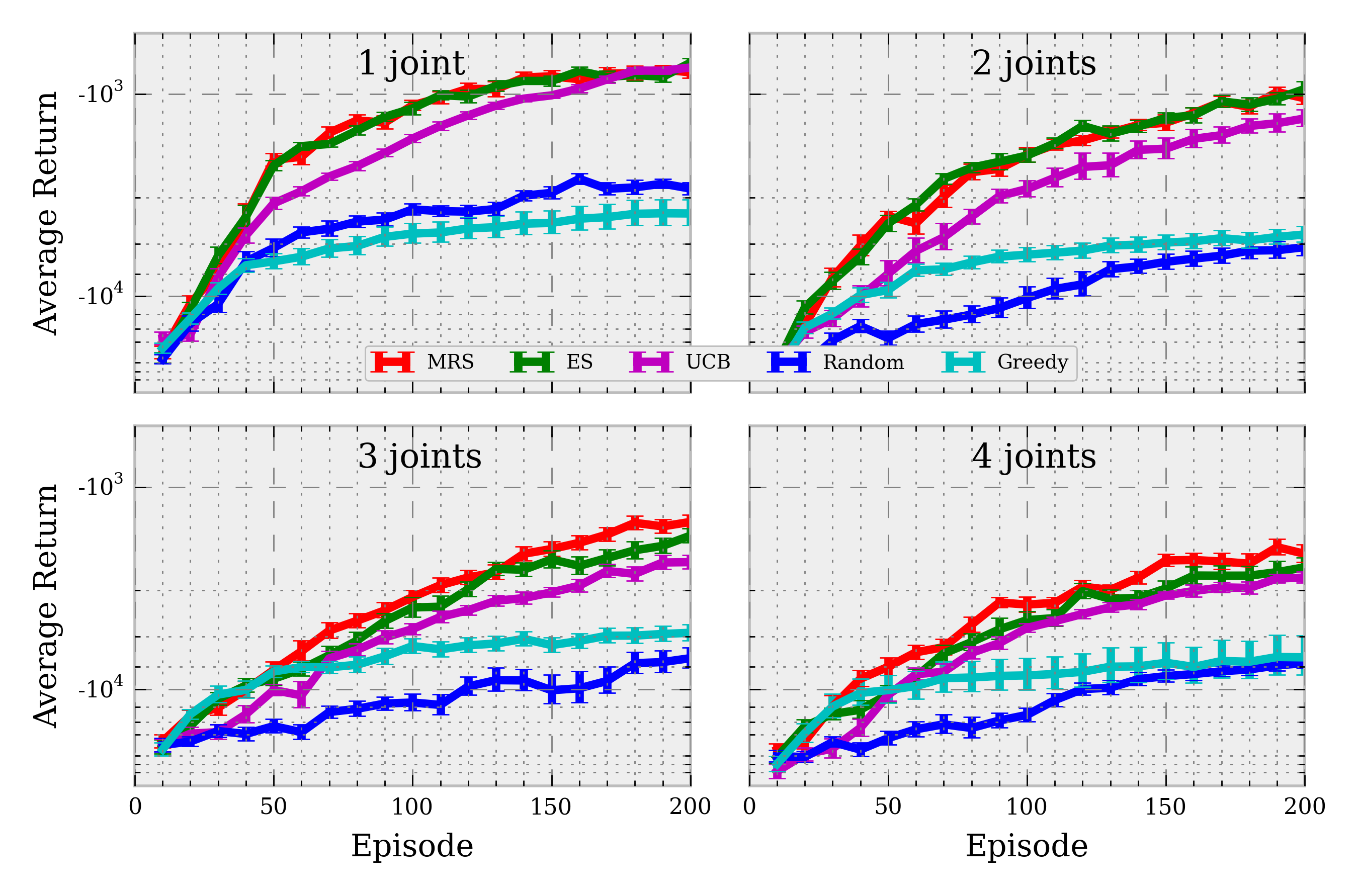}
\caption{Learning curves on the ball-throwing task for different acquisition functions and different number of controllable joints (first joint, first and second joint, and so on). Shown are mean and standard error of mean over $30$ repetitions of the greedy upper-level policy if learning would have been stopped after $N$ episodes.}
\label{fig:experiment}
\end{figure*}

Figure \ref{fig:experiment} summarizes the main results of the empirical
evaluation for different acquisition functions. Shown is the mean offline
performance of the upper-level policy at $16$ test contexts on a grid over the
context space. Selecting parameters randomly (``Random'') or greedily
(``Greedy'') during learning is shown as a baseline and indicates that
generalizing experience using a GP model alone does not suffice for quick
learning in this task. In general, performance when learning only the execution
time and the first joint is better than learning several joins at once. This is
because the execution time and first joint allow already adapting the throw to
different contexts \citep{metzen_bayesian_2015}; controlling more joints mostly
adds additional search dimensions. MRS and ES perform on par for controlling one
or two joints and outperform UCB. For higher-dimensional search spaces (three or
four controllable joints), MRS performs slightly better than ES ($p < 0.01$ after $200$ episodes for
a Wilcoxon signed-rank test). A potential reason for this might be the increasing number of areas with potentially high regret in higher dimensional spaces that may remain unexplored by ES; however, this hypothesis requires further investigation in the future.

\section{DISCUSSION AND CONCLUSION}

We have proposed MRS, a new class of acquisition functions for single- and multi-task
Bayesian optimization that is based on the principle of minimizing the expected simple
regret. We have compared MRS empirically to other acquisition functions on a
synthetic single-task optimization problem and a simulated multi-task robotic
control problem.  The results indicate that MRS performs favorably compared to the
other approaches and incurs less often a high simple regret than ES since its
objective is explicitly focused on minimizing the regret. An empirical
comparison with PES remains future work; since PES uses the same objective as ES
(minimizing $H(\mathbf{x}^\star)$), it will likely show the same deficit of
ignoring areas that have small probability $p^\star$ but could nevertheless cause a large
potential regret. On the other hand, in contrast to ES and MRS, PES allows a formal treatment
of GP hyperparameters, which can make it more sample-efficient. Potential future research on approaches for addressing
GP hyperparameters and more efficient approximation techniques for MRS would thus be desirable. Additionally,
combining MRS with active learning as done for entropy search by
\citet{metzen_active_2015} would be interesting. Moreover, we consider MRS to
be a valuable addition to the set of base strategies in a portfolio-based BO
approach \cite{shahriari_entropy_2014}. On a more theoretical level, it would be interesting if formal regret bounds can be proven for MRS.

\paragraph{Acknowledgments}
This work was supported through two grants of the German Federal Ministry of Economics and Technology (BMWi, FKZ 50 RA 1216 and FKZ 50 RA 1217).

\bibliographystyle{icml2016}
\bibliography{literature}

\newpage
\appendix 

\section{Synthetic Single-Task Benchmark with Model Mismatch}
\label{section:exp_model_mismatch}

\begin{figure*}
\centering
\includegraphics[width=.9\textwidth]{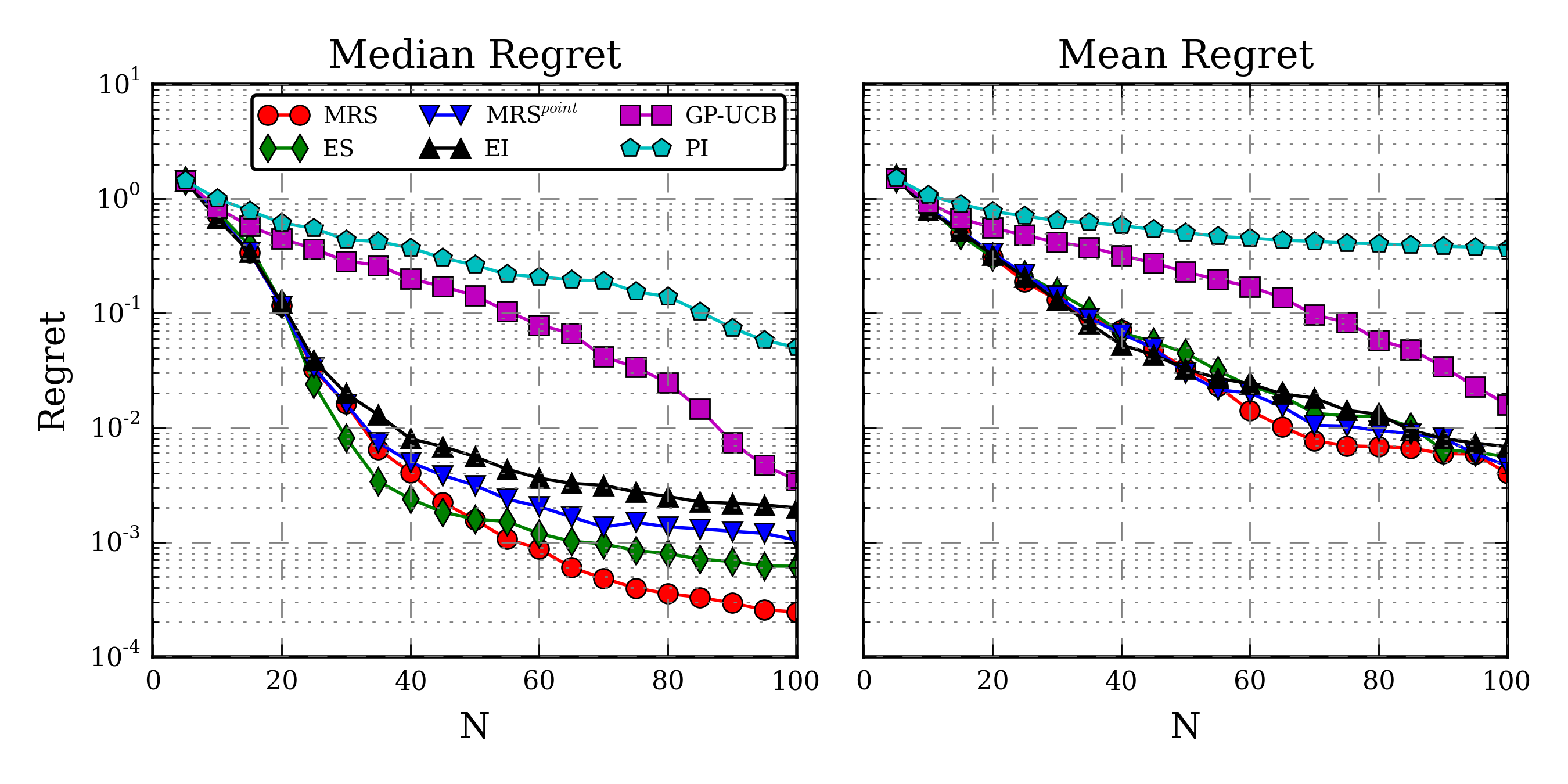}
\includegraphics[width=.9\textwidth]{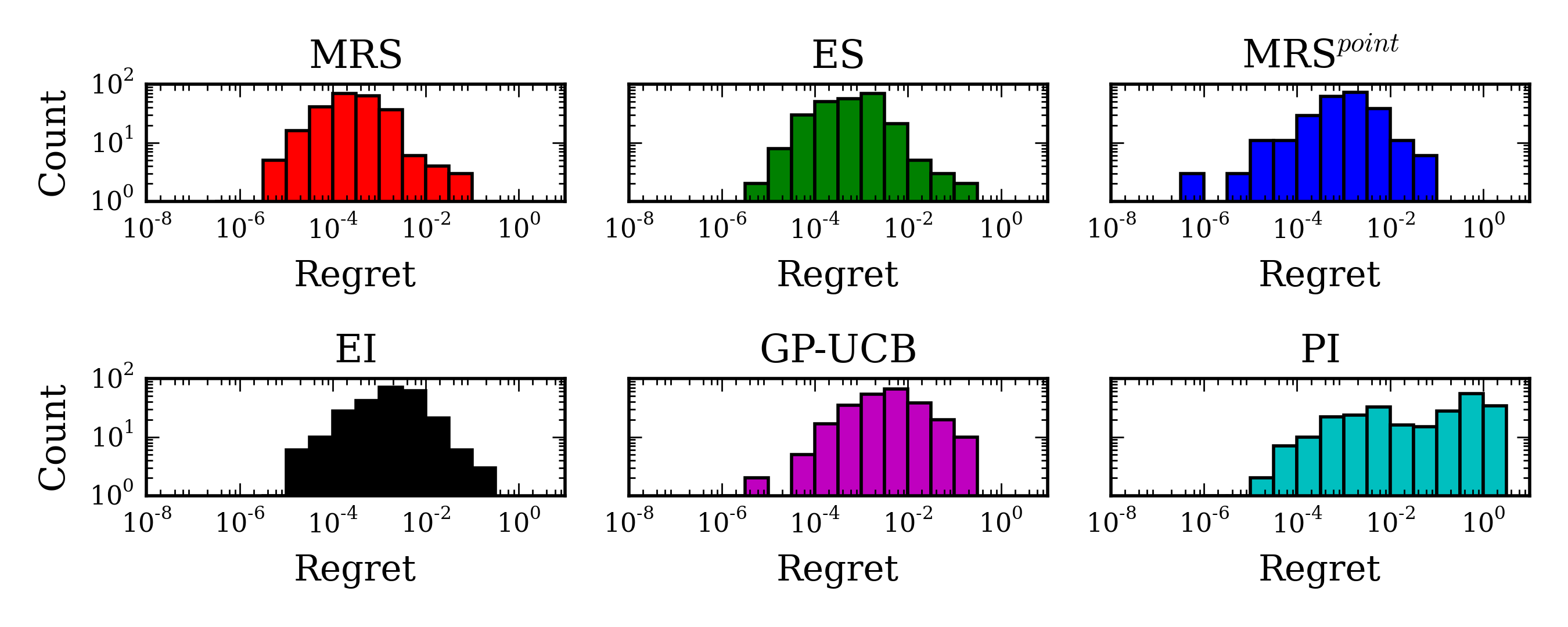}
\caption{(Top) Median and mean simple regret over $250$ repetitions for different acquisition functions. Shown is the simple regret of the recommendation $\mathbf{\tilde x}_N$ after $N$ trials, i.e., the point which maximizes the GP posterior
mean. (Bottom) Histogram of the simple regret after performing $N=100$ trials for different acquisition functions (note the log-scales).}
\label{fig:empirical_comparison_mm}
\end{figure*}

We present results for an identical setup as reported in Section \ref{Section:ResultsSingleTask},
with the
only difference being that the test functions have been sampled from a GP with
rational quadratic kernel with length scale $l=0.1$ and scale mixture
$\alpha=1.0$. The kernel used in the GP surrogate model is not modified, i.e.,
an RBF kernel with length scale $l=0.1$ is used. Thus, since different kind of
kernel govern test functions and surrogate model, we have model mismatch as
would be the common case on real-world problems. Figure
\ref{fig:empirical_comparison_mm} summarizes the results of the experiment.
Interestingly, in contrast to the experiment without model mismatch, for this setup
there are also considerable differences in the mean simple regret between MRS
and ES: while ES performs slightly better initially, it is outperformed by MRS
for $N > 60$. We suspect that this is because ES tends to explore more locally
than MRS once $p^\star$ has mostly settled onto one region of the search
space. More local exploration, however, can be detrimental in the case of
model-mismatch since the surrogate model is more likely to underestimate the
function value in regions which have not been sampled. Thus a more homogeneous
sampling of the search space as done by the more global exploration of MRS is
beneficial. As a second observation, in contrast to a no-model-mismatch
scenario, MRS$^\text{point}$ performs considerably worse than MRS when there
is model-mismatch. This emphasizes the importance of accounting for
uncertainty, particularly when there is model mis-specification.

According to the median simple regret, the difference between MRS,
MRS$^\text{point}$, ES, and EI is less pronounced in Figure
\ref{fig:empirical_comparison_mm}. Moreover, the histograms of the regret
distribution exhibit less outliers (regardless of the method). We suspect that
this stems from properties of the test functions that are sampled from a GP
with rational quadratic rather than from the model-mismatch. However,
a conclusive answer on this would require further experiments.

\end{document}